\documentclass[runningheads]{llncs}

\usepackage{eccv}

\usepackage{eccvabbrv}

\usepackage{graphicx}
\usepackage{booktabs}
\usepackage{multirow}
\usepackage{dsfont}

\usepackage[accsupp]{axessibility}  
\newcommand\norm[1]{\left\lVert#1\right\rVert}
\newcommand{\gray}[1]{\textcolor[rgb]{0.502,0.502,0.502}{#1}}

\usepackage{hyperref}

\usepackage{orcidlink}

\begin{document}

\title{Continuous Memory Representation for Anomaly Detection} 

\titlerunning{CRAD}

    
    
\author{Joo Chan Lee\inst{1}\textsuperscript{*}\orcidlink{0000-0001-9398-9089} \and
Taejune Kim\inst{1,2}\textsuperscript{*}\orcidlink{0000-0002-7581-5402} \and
Eunbyung Park\inst{1}\textsuperscript{†}\orcidlink{0000-0003-4071-2814} \and
Simon S. Woo\inst{1}\textsuperscript{†}\orcidlink{0000-0002-8983-1542} \and
Jong Hwan Ko\inst{1}\textsuperscript{†}\orcidlink{0000-0003-4434-4318}
}

\authorrunning{Lee et al.}

\institute{Sungkyunkwan University \and Robotics Lab, Hyundai Motor Company} 

\maketitle
\let\thefootnote\relax\footnotetext{\textsuperscript{*} Equal contribution.\\\textsuperscript{†} Corresponding authors.}
\begin{abstract}
There have been significant advancements in anomaly detection in an unsupervised manner, where only normal images are available for training. Several recent methods aim to detect anomalies based on a memory, comparing or reconstructing the input with directly stored normal features (or trained features with normal images).
However, such memory-based approaches operate on a discrete feature space implemented by the nearest neighbor or attention mechanism, suffering from poor generalization or an identity shortcut issue outputting the same as input, respectively.
Furthermore, the majority of existing methods are designed to detect single-class anomalies, resulting in unsatisfactory performance when presented with multiple classes of objects.
To tackle all of the above challenges, we propose CRAD, a novel anomaly detection method for representing normal features within a ``continuous'' memory, enabled by transforming spatial features into coordinates and mapping them to continuous grids. 
Furthermore, we carefully design the grids tailored for anomaly detection, representing both local and global normal features and fusing them effectively.
Our extensive experiments demonstrate that CRAD successfully generalizes the normal features and mitigates the identity shortcut, furthermore, CRAD effectively handles diverse classes in a single model thanks to the high-granularity continuous representation.
In an evaluation using the MVTec AD dataset, CRAD significantly outperforms the previous state-of-the-art method by reducing 65.0\% of the error for multi-class unified anomaly detection. Our project page is available at \href{https://tae-mo.github.io/crad/}{https://tae-mo.github.io/crad/}.
\end{abstract}

\begin{figure*}[t]
    \begin{center}
    \includegraphics[width=1.0\linewidth]{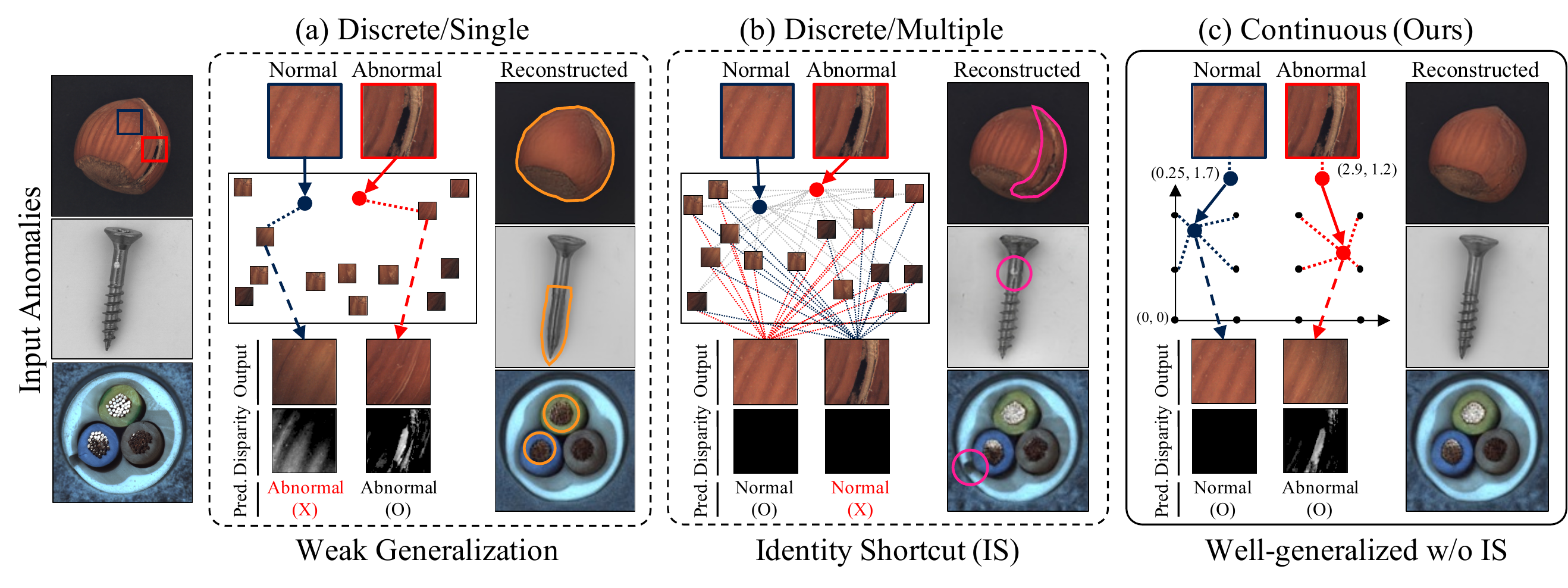}
    \end{center}
    \caption{Conceptual diagram and qualitative results of existing methods and ours. (a) and (b) use single and multiple normal features in a discrete memory, respectively, while our method (c) exploits continuous feature memory. We visualize the anomaly detection process with the normal (navy) and abnormal (red) patches of the top-left reference image. `Pred.' indicates the prediction based on the disparity, and wrong predictions are marked as (X) with red color. We present the reconstruction results based on the reference abnormal images.}
\label{fig:space}
\end{figure*}

\section{Introduction}
With the recent advances in deep neural networks, anomaly detection (AD) has been applied for a wide range of applications such as manufacturing industry~\cite{indu_survey,patchcore}, video surveillance~\cite{surveil_survey, surveil}, and medical imaging~\cite{schlegl2019f, squid}.
Despite its success, there are still several limitations hindering the broader applicability across many practical scenarios.
In particular, one major bottleneck is the collection of a sufficient amount of anomalous data, which is scarce by definition of an anomaly. Furthermore, many AD systems often require considerable effort for pixel-wise labeling of `normal' versus `abnormal' data.
Due to these challenges, there has been a growing interest in developing methods in an unsupervised manner where we train AD models solely with normal data~\cite{patchcore}.

As notable examples, PatchCore~\cite{patchcore}, PaDiM~\cite{padim}, and SPADE\cite{spade} proposed using additional memory that directly stores normal features (or distributions). 
During inference, these memory-based methods detect anomalies based on the distance between the testing input and its nearest neighbor in their memory, as shown in~\cref{fig:space}(a). 
While showing promising performance, these methods require storing a wide array of diverse normal features in memory, resulting in high space complexity and resource-intensive search operations.
Moreover, these methods are often ineffective in identifying the characteristics of global anomalies due to their diversity, leading to suboptimal AD performance (see~\cref{tab:space}).

Another line of work~\cite{gong2019memorizing, park2020learning, hou2021divide} focuses on producing generalized normal features.
Unlike the aforementioned approaches that use the nearest neighbor technique, these methods combine multiple normal features from the memory using an attention mechanism (i.e., referring to multiple discrete features), given a normal or abnormal input (\cref{fig:space}(b)). 
They assume that the model always generates normal features, regardless of whether the inputs are normal or abnormal, thus anomalous regions can be detected based on the disparity between the inputs and outputs.
Because these models gather diverse normal features from the memory via attention, they have exhibited increased robustness to test data, leading to improved generalization performance.
However, such strength may turn into a drawback when testing abnormal inputs. 
If these abnormal inputs can be reconstructed using a combination of normal features, the model could potentially generate outputs that are identical to the abnormal inputs. 
This issue, referred to as an identity shortcut (IS) by UniAD~\cite{uniad}, prevents the models from detecting anomalies due to the minimal difference between the abnormal input and produced output.

Furthermore, a significant limitation exists in most of the approaches discussed earlier, as they are primarily designed to handle only one class of objects per model.
When these methods are extended to multi-class scenarios, where a single model handles multiple classes, a significant performance drop has been observed~\cite{uniad}, even with state-of-the-art methods.
To mitigate this limitation, memory-based methods may incorporate sufficient memory to accommodate multi-class normality, yet this simultaneously increases memory consumption and search latency.
Moreover, attention-based methods experience more severe challenges with the IS problem, as the greater number and diversity of aggregated features tend to more easily represent anomalies.
These challenges necessitate training distinct models for each class, which increases training complexity, memory usage, computational overheads, and even data preparation efforts in a practical implementation.

To address all of the issues above, we propose a novel continuous memory representation for anomaly detection (CRAD), where we use an external grid-based representation for normal features (\cref{fig:space}(c)).
Given that the input to the grid is a spatial feature from an image rather than coordinates required for conventional grids, we need a specially designed framework for handling feature-based inputs.
In light of this requirement, we transform input spatial features into low-dimensional coordinates, based on which we interpolate neighboring normal features in the grid.
This continuous memory, unlike other discrete counterparts, allows for instant ($\mathcal{O}(1)$ time complexity) retrieval of the normal feature, while the interpolation technique mitigates the weak generalization issue.
Also, as our approach does not rely on innumerable features across the entire memory, it reduces the risk of generating entirely new features (unseen anomalies in our context), thus helping to avoid the IS problem.
These advantages are even stronger in the multi-class scenario, where a larger memory space would be needed, by avoiding the tremendous computation associated with searching or aggregating every feature.

Deploying the continuous memory, we additionally include specific designs on CRAD tailored for AD.
CRAD incorporates two distinct continuous memories to represent normal features from both a local and global perspective. 
Through the integration of these representations, CRAD adeptly identifies coarse-to-fine anomalies. 
Furthermore, we implement coordinate jittering to enhance the generalization capability of the grid, facilitating the update of a broader range of grid values with each input coordinate. 
A feature refinement process further improves CRAD, minimizing false detections (i.e., false positives) by ensuring that the normal regions in the fused output remain consistent with the original input.

CRAD not only addresses weak generalization and IS issues but also provides several additional advantages.
Whereas discrete memories struggle to represent global features (e.g., an entire input feature), our continuous memory successfully captures their structural characteristics (\cref{tab:space}). 
Therefore, it enables identifying anomalies across a wider range of classes, each featured by distinct structures.
Furthermore, different from discrete memories with limited entries, CRAD represents an infinite number of normal features in the continuous memory.
Thus, CRAD achieves high performance with compact memory, resulting in high parameter efficiency.
To the best of our knowledge, this is the first work leveraging continuous memory to effectively represent normal features for AD.

In the extensive experiments, we demonstrate the superiority of continuous memory over the existing discrete spaces in terms of accuracy and efficiency (both for computation and parameter).
Furthermore, experimental results on the MVTec AD dataset~\cite{mvtec} show that CRAD achieves state-of-the-art performance in a unified setting (multi-class AD with a single model) for anomaly detection, which even outperforms the state-of-the-art models trained for each respective class.
With the comprehensive analysis, we demonstrate CRAD is an effective solution for AD, overcoming the limitations of existing methods. 

\section{Related Work}
\label{sec:rw}
\subsection{Unsupervised Anomaly Detection}
Confronted with the difficulties in collecting and annotating anomalous data, recent works have focused on an unsupervised approach, where only normal images are available for training.
Several studies have explored how a model trained only on normal data behaves differently when exposed to anomalous test inputs.
For instance, reconstruction-based methods~\cite{zhou2020encoding,liang2022omni,uniad,ristea2022self} utilize auto-encoders~\cite{bergmann2018improving,baur2019deep,ye2020attribute} or GANs~\cite{akcay2019ganomaly,schlegl2019f,pidhorskyi2018generative,sabokrou2018adversarially} to reconstruct the normal feature regardless of input's normality, and then compare the reconstructed outcomes and original inputs to detect and localize the anomalies.
Similarly, distillation-based methods~\cite{salehi2021multiresolution,rd4ad,bergmann2020uninformed,wang2021student} exploit the disparity between the output of student and teacher networks on anomalous input. 

Other methods leverage auxiliary memory to retain normal features, where they are classified into the following two categories: reference to a single discrete feature (\cref{fig:space}(a)) and reference to a combination of discrete features (\cref{fig:space}(b)).
The former methods~\cite{spade,rippel2021modeling,padim,patchcore} detect anomalies by measuring the distance between the input and stored features (or feature distributions) extracted by a pre-trained network.
For instance, PatchCore~\cite{patchcore} and SPADE~\cite{spade} are designed to store the representative normal features in a memory bank and use the nearest neighbor search for anomaly scoring.
However, the above methods present a significant challenge to represent features that are not already stored in their memory.
Therefore, when faced with a complex and diverse range of inputs, they result in limited performance or require tremendous memory usage to achieve satisfactory performance.
Moreover, an increase in the number of stored features can significantly delay the time to reference all the stored ones, further hampering their effectiveness.

The latter methods~\cite{gong2019memorizing,hou2021divide,park2020learning} employ attention-like techniques to take a weighted sum of all normal features in the discrete space based on their similarity to the input. 
While these approaches exhibit superior generalization capabilities compared to the former methods, they generalize not only to normal features but also to abnormal features, using combinations of all features in the memory.
This causes the input anomalies to be reconstructed, coined as the IS problem, reducing the AD performance by hindering the model from recognizing the disparity, as depicted in~\cref{fig:space}(b).

\subsection{Unified Model for Multiple Classes}
While the approaches mentioned above exhibit promising performance in identifying anomalies within a single class, they might not be easily and practically deployable due to various issues.
When targeting multi-class objects, the number of required models increases, resulting in multiplied memory and computational overhead. 
Moreover, training numerous models in proportion to the number of object classes further complicates their practical implementation.
Conversely, when these methods are applied to address multiple classes with a single model to avoid the above issues, they suffer from a significant performance drop~\cite{uniad}.
This is because multi-class data pose a more complex problem for models originally designed for a single class, where more classes entail more complex and diverse underlying class distributions.

Recently, UniAD~\cite{uniad} introduced a framework capable of detecting and localizing anomalies in multi-classes setting with a single model. 
UniAD defines the IS problem, which means the reconstruction-based models tend to be trained as an identity function, thereby outputting the same as input even if the input contains anomalies.
This hinders the model from identifying anomalies based on the disparity between the input and output.
To mitigate the IS issue, UniAD introduces a learnable query with neighbor-masked attention (NMA). 
NMA restricts each query feature from attending an input feature in the same and neighboring location.
However, UniAD shows limited performance due to the lack of a special design for multi-class scenarios, such as employing fixed queries regardless of the input's class or visual characteristics.
Although several recent works have explored on unified AD framework by using synthesized anomalies~\cite{omnial} and vector quantization~\cite{hvq}, they still show limited detection performance.

\subsection{Grid Representation}
In the revolution of neural fields or neural representations that parameterize signals by a function of coordinates, grid representation has been demonstrated to be effective in various tasks, including image and video processing~\cite{img_grid, ffnerv}, 3D reconstruction~\cite{occupancy,loc_grid_3dscene}, and novel view synthesis~\cite{plenoxels,tensorf,nsvf}. The grid structure is capable of efficiently representing high-frequency components without spectral bias~\cite{spectral, cam}, and effectively generalizing features by offering a continuous feature space.

In this work, we propose incorporating grid representation to achieve high-performance AD.
Our key contribution involves representing the normal features in a continuous space by substituting the discrete feature memory to the continuous grid in order to resolve the challenging issues discussed above while achieving high performance.

\begin{figure*}[t]
    \begin{center}
    \includegraphics[width=1.0\linewidth]{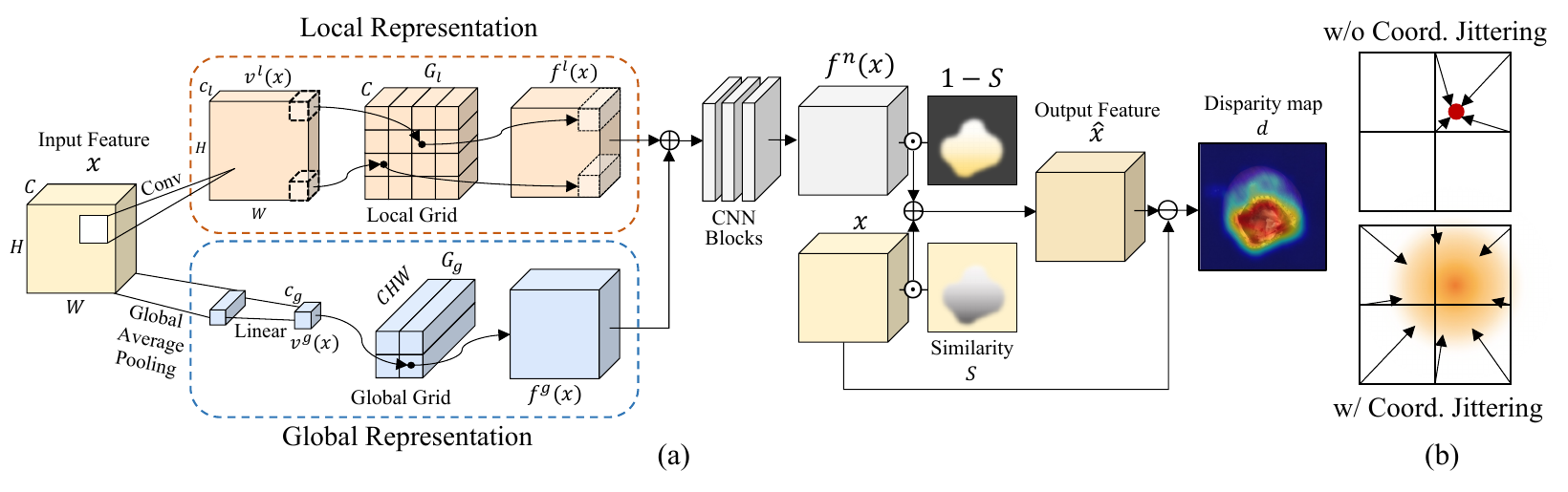}
    \end{center}
    \caption{(a) The detailed architecture of CRAD and (b) visualization of coordinate jittering. The input $x$ is firstly transformed into pixel-wise and feature-wise coordinates. After the normal features are sampled from local and global representations, they are fused by CNN blocks. The final reconstruction is acquired through the proposed feature refinement process.}
\label{fig:detail}
\end{figure*}
\section{CRAD}
\noindent\textbf{Background. }
To help the readers understand CRAD, we first describe the grid operation. A grid is trained as a function of coordinates with infinite resolution, outputting coordinate-corresponding features.
The output feature in infinite resolution is aggregated by nearby features in the grid, based on the distance between the coordinate of the input and neighboring features.
For example, when we take 1D grid sampling $\phi(\cdot;G):\mathbb{R} \rightarrow \mathbb{R}^C$, the output feature with channel $C$ is interpolated by neighboring values of the 1D grid $G\in \mathbb{R}^{R\times C}$, which is mathematically formulated as follows:
\begin{equation}
\label{bi_intp}
\begin{gathered}
    \phi(v;G)=|v-n|G[m]+|v-m|G[n],\\
    m=\lfloor v\rfloor,\, n=\lceil v \rceil,
\end{gathered}
\end{equation}
where $v\in \mathbb{R}$ is an arbitrary input coordinate normalized to the grid resolution $R$, and $G[i]$ denotes the feature from index $i$ of the grid $G$.
$m$ and $n$ are indices to be referenced, and $\lfloor\cdot\rfloor$ and $\lceil\cdot\rceil$ denote floor and ceiling operation, respectively.
The above equation can be simply extended to a higher dimension $D$ by interpolating $2^D$ values of a $D$-dimensional grid (e.g., $2^D$ = 4 values in a 2D grid in~\cref{fig:space}(c)).

\noindent\textbf{Overview. } The motivation of our work is to effectively represent the normal features in continuous memory using the grid operation, distinct from the discrete memories.
In an unsupervised manner, CRAD detects anomalous images and regions based on the discrepancy between the input feature and output normal feature, as described in~\cref{fig:detail}(a).
Therefore, the primary objective of CRAD is to effectively retain the normal components (e.g., shapes or textures) of the original input feature while eliminating any anomalies presented within the feature. 

To this end, we represent normal features in the continuous memory during the training phase, coined as normal representation, which is used for replacing abnormal features in the testing phase. 
We describe how CRAD represents normal features in the continuous memory and acquires the output feature $\hat{x}$, based on the input feature $x$ extracted from a pre-trained backbone, where $x$, $\hat{x} \in \mathbb{R}^{C\times H\times W}$, and $C,H,W$ are the channel, height, and width of the feature, respectively.

\subsection{Normal Representation}
The fundamental concept of CRAD is to transform the input feature into specific coordinates of continuous values, which are subsequently mapped to feature grids.
In particular, we design to represent the normal features from local and global perspectives.
By combining the distinctive features from each perspective, the resulting feature can provide a strong representation of the input, capturing both fine-grained details as well as broader overall structures.

\noindent\textbf{Local representation.}
As shown in~\cref{fig:detail}(a), CRAD samples each pixel of the feature, which characterizes each patch of the image, to represent the local feature.
Then, the channels of each pixel are transformed to corresponding coordinates (a low-dimensional vector) by convolutional layers with a kernel size of 1, followed by hyperbolic tangent activation.
Using these pixel-wise coordinates, we obtain normal features sampled from the local grid representation.
More formally, we define a function $v^l(\cdot) : \mathbb{R}^{C\times H\times W} \rightarrow \mathbb{R}^{C_l\times H\times W}$, which generates pixel-wise coordinates based on the input feature, where $C_l$ is the dimension of the produced coordinates.
Given the pixel-wise coordinates $v^l_{h,w}(\cdot)\in\mathbb{R}^{C_l}$, normal features are sampled from $C_l$-dimensional grid $G_l$, which has the resolution of each dimension $R_l$ and channel of $C$.
The equation for local representation $f^l(x):\mathbb{R}^{C\times H\times W} \rightarrow \mathbb{R}^{C\times H\times W}$ is written as follows:
\begin{gather}\label{eq:local}
    f^l_{h,w}(x) = \phi(v^l_{h,w}(x); G_{l}),
\end{gather}
where $\phi(\cdot;G_l): \mathbb{R}^{C_l} \rightarrow \mathbb{R}^{C}$ represents sampling feature from grid $G_l$ by bilinearly interpolating the grid values based on the coordinates.

As each pixel of the feature characterizes a patch in an image, the local representation ensures retaining normal patches and replacing abnormal patches with normal patches that have similar local context.
Hence, when a normal patch is fed, even though there is no exact match in the training patches, a corresponding normal feature can be represented by interpolating normal features mapped at nearby coordinates.

In addition, for an abnormal patch, CRAD finds a normal feature that is the most representative of the abnormal patch based on the reduced coordinates.
As the grid has never been exposed to abnormal features during training, it is unable to represent abnormal features by interpolating nearby normal features.
This is the core idea of how we can effectively resolve the identity shortcut (IS) issue frequently found in the existing methods that aggregate numerous features based on similarities using attention mechanisms~\cite{gong2019memorizing, park2020learning}.

\noindent\textbf{Global representation.}
Anomalous regions can exist not only locally within an image but also at a global scale.
To handle such global anomalous cases, CRAD maintains another grid representation to capture the global feature of an image.
Similar to the local representation, we formulate the function to obtain global feature coordinates $v^g(\cdot) : \mathbb{R}^{C\times H\times W} \rightarrow \mathbb{R}^{C_g}$, where $C_g$ is the reduced dimension of coordinates.
For the function $v^g(\cdot)$, we employ global average pooling and linear layers, as shown in~\cref{fig:detail}(a).
The feature-wise coordinates are mapped to each normal feature by $C_g$-dimensional grid $G_g$ that has the resolution of each dimension $R_g$. An element of the grid $G_g$ is a $CHW$ dimensional vector, which is reshaped to $C \times H \times W$ tensor once sampled.
The equation for global representation $f^g(x):\mathbb{R}^{C\times H\times W} \rightarrow \mathbb{R}^{C\times H\times W}$ is expressed as follows:
\begin{gather}\label{eq:global}
    f^{g}(x) = \mathsf{reshape}(\phi(v^{g}(x); G_{g})),
\end{gather}
where $\phi(\cdot;G_g) : \mathbb{R}^{C_g} \rightarrow \mathbb{R}^{CHW}$ represents sampling feature from grid $G_g$ by bilinear interpolation, and $\mathsf{reshape}(\cdot) : \mathbb{R}^{CHW} \rightarrow \mathbb{R}^{C\times H\times W}$ denotes the reshape operation.

The global representation not only effectively replaces global anomalies as a whole but also distinguishes the class-wise distribution for the unified setting.
Based on the image-wise features, the reduced coordinates are well distributed on the continuous space, modeling the decision boundary of complex distribution (see~\cref{fig:coord}(b) and~\cref{sec:eff} for more information).

\noindent\textbf{Fused representation.}
We combine the local and global representations $f^l(x)$ and $f^g(x)$ to effectively learn the normal representation $f^n(x)$, as shown in~\cref{fig:detail}(a).
The local and global representations are concatenated and then fed into the following convolution networks $\psi(\cdot) : \mathbb{R}^{2C \times H \times W} \rightarrow \mathbb{R}^{C \times H \times W}$ to reconstruct $f^n(x)$ as follows:
\begin{gather}
    f^{n}(x) = \psi(\mathsf{concat}(f^{l}(x), f^{g}(x))),
\end{gather}
where $\mathsf{concat}(\cdot, \cdot)$ denotes the concatenation of two features along with the channel axis.
By fusing the local and global representation, CRAD can represent normal features from fine-grained details to broader contexts, resulting in higher performance compared to the cases using only either of them (see ablation study in~\cref{sec:ablation}). 

\subsection{Feature Refinement}\label{feature_refinement}
Despite the fusion of local and global normal representation, deviations for the normal regions between $f^n(x)$ and $x$ can still exist, which can lead to false detection (i.e., false positives).
Hence, in feature refinement, we aim to refine $f^n(x)$ in the regions that are supposed to be normal but deviate from $x$, with the goal of reducing false positives.
To identify such regions, we evaluate the pixel-wise similarity between $x$ and $f^n(x)$ by combining both Mean Squared Error (MSE) and cosine similarity.
These two metrics offer a comprehensive view of the differences between normal and abnormal features, where MSE captures the absolute intensity disparities while cosine similarity characterizes structural and positional similarity.
By considering the combined similarity $S\in\mathbb{R}^{H\times W}$, we can reconstruct $\hat{x}$ as follows:
\begin{gather}
    \hat{x}_{h,w} = S_{h,w} x_{h,w} + (1-S_{h,w}) f^n_{h,w}(x), \\
    S_{h,w}=\lambda_1 \mathds{1}[\mathsf{mse}(x_{h,w}, f_{h,w}^n(x)) < k] + \lambda_2 \mathsf{cosim}(x,f^n(x)),
\end{gather}
where $h,w$ are the indices of the spatial feature, $\mathds{1}[\cdot]$ is the indicator function and $\mathsf{mse}(\cdot,\cdot)$ and $\mathsf{cosim}(\cdot,\cdot)$ are the MSE and cosine similarity, respectively.
To use MSE as a measure of similarity, we convert the MSE value to either 0 or 1, depending on whether it surpasses the threshold $k$ or not.

\subsection{Training and Inference}
\noindent\textbf{Coordinate jittering.}
To achieve a more generalized grid representation, we apply Gaussian noise to vectorized local coordinates $v^l(x)$ in the training phase.
For instance, without jittering, a coordinate affects up to four grid values in a 2D grid, as shown in~\cref{fig:detail}(b).
In contrast, when perturbating the coordinate, we can update more grid values with bell-shaped distribution in each iteration, producing a more generalized grid.

\noindent\textbf{Training.}
Given $x$ and $\hat{x}$ derived from CRAD, we employ the MSE loss as an objective function, as follows: 
\begin{gather}
    \mathcal{L} = \frac{1}{CHW}\norm{x-\hat{x}}_2^2.
\label{eq:mseloss}
\end{gather}

\noindent Based on \cref{eq:mseloss}, we learn the entire model in an end-to-end manner, including the grids initialized by Xavier normal initialization~\cite{xavier}. As $x$ is always a normal input in the training phase, the grids are learned to represent normal features.

\noindent\textbf{Inference.}
To perform anomaly detection and localization through the disparity between $x$ and $\hat{x}$, an anomaly score map $d \in \mathbb{R}^{H\times W}$ is formulated as follows:
\begin{gather}
    d_{h,w} = \norm{x_{h,w}-\hat{x}_{h,w}}_2, 
\end{gather}

where $h$ and $w$ indicate the location of each pixel.
To match with the corresponding ground truth, $d$ is interpolated into the original shape of the input. An anomaly score for each image is obtained by taking the max value from the average-pooled $d$, and the interpolated anomaly map itself is used for the pixel-wise anomaly score.

\section{Experimental Results} \label{sec:experiment}
\subsection{Experimental Setup}
We used MVTec AD~\cite{mvtec} and VisA~\cite{visa} datasets, which are representative datasets for real-world unsupervised AD.
We evaluated the performance of anomaly detection by the Area Under the Receiver Operator Curve (AUROC). 
Following previous studies, we computed the class-average AUROC for detection and pixel-wise AUROC for localization.
We implemented CRAD in the PyTorch framework, and we used the NVIDIA A5000 GPU for all evaluations.
We trained our models for 50 epochs, thrice with different seeds (0,1,2), with a batch size of 64.
We describe the detailed implementation of CRAD in the supplementary materials.

We evaluated the performance under two different scenarios: 1) a unified setting where a single model is used for anomaly detection across multiple classes, and 2) a separate setting in which we utilize respective models for different classes.
When training a unified model across all methodologies, we maintained the model size to be consistent with each separate model.

\subsection{Effectiveness of Continuous Memory Representation}
\label{sec:eff}
\begin{minipage}[b]{0.578\textwidth} \noindent \textbf{Improved performance.} 
To assess the efficacy of the continuous memory, we implemented two baselines with discrete memories under the same overall detection framework of CRAD as follows: 1) referring to a single feature from discrete space (\cref{fig:space}(a)) through vector quantization (VQ), and 2) referring to a combination of multiple discrete features (\cref{fig:space}(b)) with an attention module.
As shown in~\cref{tab:space}, CRAD, providing a continuous memory, outperforms the other baselines for both local and global representation.
When we expand the memory size for local representation, the attention shows the performance drop, suffering from a more severe IS.
\end{minipage}\hfill
\begin{minipage}[b]{0.40\textwidth}
\captionof{table}{Performance evaluation of the different feature memories in the unified setting. \#Entry denotes the number of features in each memory and Persp. indicates the perspective (local or global).}
\centering
\resizebox{1.0\textwidth}{!}{
\begin{tabular}{@{}ccccc@{}} 
\toprule
Persp. & Method & \#Entry & Detection & Localization \\ 
\midrule
\multirow{5}{*}[-0.5em]{Local} & \multirow{2}{*}{VQ} & 64 & 96.9±0.65 & 96.1±0.05 \\
 &  & 256 & 97.8±0.23 & 96.0±0.08 \\ 
\cmidrule(lr){2-3}\cmidrule(lr){4-5}
 & \multirow{2}{*}{Attention} & 64 & 95.9±1.1 & 96.2±0.25 \\
 &  & 256 & 93.9±1.8 & 95.1±0.76 \\ 
\cmidrule(lr){2-3}\cmidrule(lr){4-5}
 & CRAD & 64 & \textbf{98.6±0.07} & \textbf{97.5±0.04} \\ 
\midrule
\multirow{5}{*}[-0.5em]{Global} & \multirow{2}{*}{VQ} & 16 & 81.0±0.97 & 89.9±0.42 \\
 &  & 64 & 82.2±0.56 & 91.4±1.0 \\ 
\cmidrule(lr){2-3}\cmidrule(lr){4-5}
 & \multirow{2}{*}{Attention} & 16 & 77.9±2.3 & 86.7±3.4 \\
 &  & 64 & 82.1±0.54 & 90.6±0.19 \\ 
\cmidrule(lr){2-3}\cmidrule(lr){4-5}
 & CRAD & 16 & \textbf{92.3±0.60} & \textbf{95.7±0.17} \\
\bottomrule
\end{tabular}}
\label{tab:space}
\end{minipage}
Although VQ shows performance improvement with larger memory entries, it still falls short of CRAD even with the quadrupled feature space.
Furthermore, the baselines consistently underperform in global representation, indicating their inability to represent structural information of the entire feature.

\begin{figure}[t]
    \begin{center}
    \includegraphics[width=1.0\textwidth]{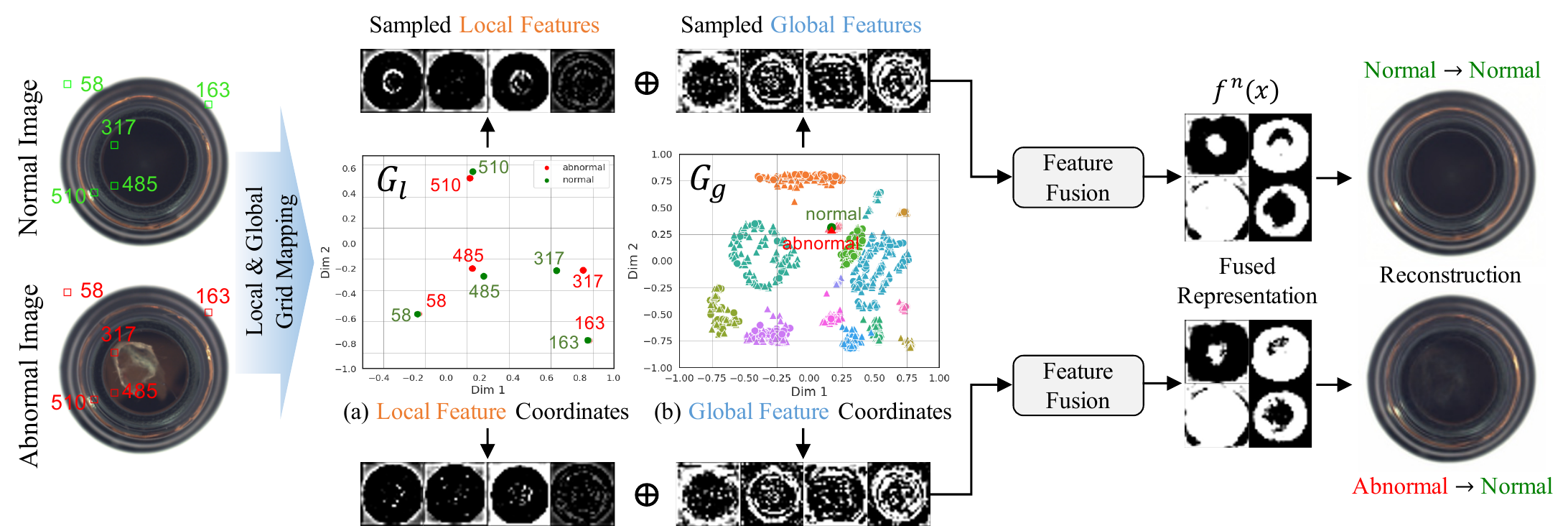}
    \end{center}
    \caption{Visualization of CRAD's pipeline. Each marker in (a) corresponds to the patch on the left image that has the same number and color. Each marker in (b) corresponds to a single image from the test dataset, where different colors represent distinct classes, and circles and triangles denote the normal and abnormal images, respectively. `Dim 1' and `Dim 2' are the two dimensions of 2D grids.}
    \label{fig:coord}
\end{figure}

\noindent \textbf{Visualization of coordinates.} Although the quantitative results above clearly demonstrate the effectiveness of the continuous normal representation of CRAD, we additionally visualize the generated and mapped coordinates in~\cref{fig:coord}.
The normal and abnormal areas with a similar local characteristic are mapped at a near distance in the local feature space (e.g., the patch $317$ and $485$ in~\cref{fig:coord}(a)).
Similarly, the global coordinates of the two input images are mapped to almost the same location at the global feature space (\cref{fig:coord}(b)).
These results indicate that the model successfully learns to generate coordinates corresponding to each input feature, and the local and global grids can represent the normal features from each perspective effectively.

\begin{figure}[h]
    \begin{center}
    \includegraphics[width=0.75\textwidth]{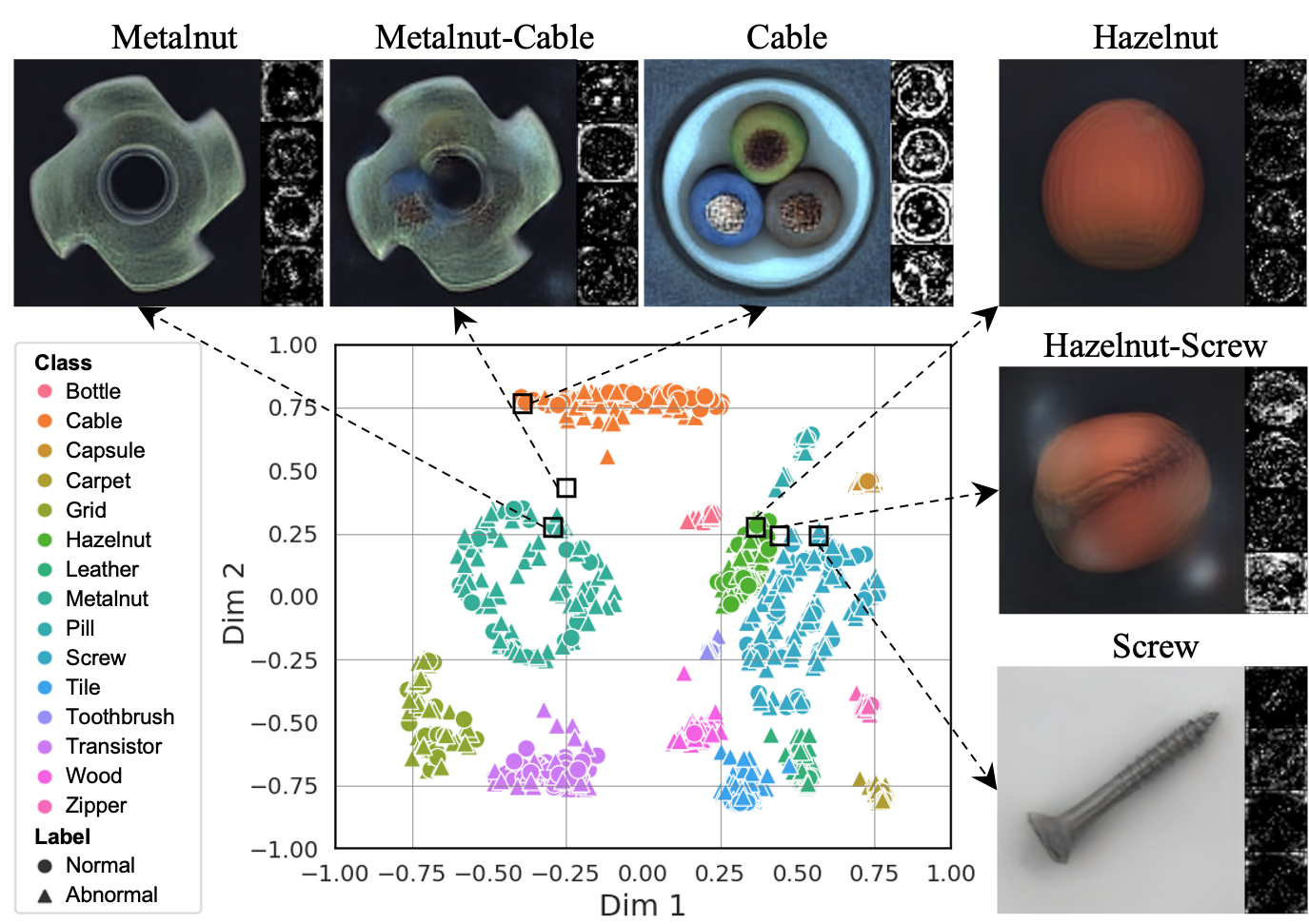}
    \end{center}
    \caption{Visualization of the contents mapped at a continuous grid. We manually select six global coordinates and visualize the corresponding sampled normal features.}
    \label{fig:global_vis}
\end{figure}

\begin{table}[h!]
\centering
\resizebox{\textwidth}{!}{%
\begin{tabular}{lccccccccc} 
\toprule
Class & US~\cite{bergmann2020uninformed} & PaDiM~\cite{padim} & MKD~\cite{salehi2021multiresolution} & DRAEM~\cite{draem} & RD4AD~\cite{rd4ad} & PatchCore~\cite{patchcore} & UniAD~\cite{uniad} & HVQ-T~\cite{hvq} & CRAD (Ours) \\ 
\midrule
Bottle & 84.0/\gray{99.0} & 97.9/\gray{99.9} & 98.7/\gray{99.4} & 97.5/\gray{99.2} & 98.7/\gray{100} & 100/\gray{100} & 99.7/\gray{100} & 100/\gray{-} & 100±0.00/\gray{100} \\
Cable & 60.0/\gray{86.2} & 70.9/\gray{92.7} & 78.2/\gray{89.2} & 57.8/\gray{91.8} & 85.0/\gray{95.0} & 99.7/\gray{99.4} & 95.2/\gray{97.6} & 99.0/\gray{-} & 99.1±0.34/\gray{99.7} \\
Capsule & 57.6/\gray{86.1} & 73.4/\gray{91.3} & 68.3/\gray{80.5} & 65.3/\gray{98.5} & 95.5/\gray{96.3} & 90.9/\gray{97.8} & 86.9/\gray{85.3} & 95.4/\gray{-} & 97.0±0.05/\gray{98.4} \\
Hazelnut & 95.8/\gray{93.1} & 85.5/\gray{92.0} & 97.1/\gray{98.4} & 93.7/\gray{100} & 87.1/\gray{99.9} & 100/\gray{100} & 99.8/\gray{99.9} & 100/\gray{-} & 100±0.06/\gray{100} \\
Metal Nut & 62.7/\gray{82.0} & 88.0/\gray{98.7} & 64.9/\gray{73.6} & 72.8/\gray{98.7} & 99.4/\gray{100} & 99.9/\gray{100} & 99.2/\gray{99.0} & 99.9/\gray{-} & 100±0.00/\gray{100} \\
Pill & 56.1/\gray{87.9} & 68.8/\gray{93.3} & 79.7/\gray{82.7} & 82.2/\gray{98.9} & 52.6/\gray{96.6} & 96.9/\gray{96.0} & 93.7/\gray{88.3} & 95.8/\gray{-} & 98.6±0.36/\gray{98.7} \\
Screw & 66.9/\gray{54.9} & 56.9/\gray{85.8} & 75.6/\gray{83.3} & 92/\gray{93.9} & 97.3/\gray{97.0} & 90.1/\gray{97.0} & 87.5/\gray{91.9} & 95.6/\gray{-} & 97.6±0.33/\gray{98.6} \\
Toothbrush & 57.8/\gray{95.3} & 95.3/\gray{96.1} & 75.3/\gray{92.2} & 90.6/\gray{100} & 99.4/\gray{99.5} & 100/\gray{99.7} & 94.2/\gray{95.0} & 93.6/\gray{-} & 99.2±0.73/\gray{96.1} \\
Transistor & 61.0/\gray{81.8} & 86.6/\gray{97.4} & 73.4/\gray{85.6} & 74.8/\gray{93.1} & 92.4/\gray{96.7} & 99.7/\gray{100} & 99.8/\gray{100} & 99.7/\gray{-} & 99.8±0.18/\gray{99.9} \\
Zipper & 78.6/\gray{91.9} & 79.7/\gray{90.3} & 87.4/\gray{93.2} & 98.8/\gray{100} & 99.6/\gray{98.5} & 94.7/\gray{99.5} & 95.8/\gray{96.7} & 97.9/\gray{-} & 99.2±0.13/\gray{99.6} \\ 
\midrule
Carpet & 86.6/\gray{91.6} & 93.8/\gray{99.8} & 69.8/\gray{79.3} & 98.0/\gray{97.0} & 97.1/\gray{98.9} & 97.1/\gray{98.7} & 99.8/\gray{99.9} & 99.9/\gray{-} & 99.9±0.05/\gray{100} \\
Grid & 69.2/\gray{81.0} & 73.9/\gray{96.7} & 83.8/\gray{78.0} & 99.3/\gray{99.9} & 99.7/\gray{100} & 96.3/\gray{97.9} & 98.2/\gray{98.5} & 97.0/\gray{-} & 100±0.0/\gray{100} \\
Leather & 97.2/\gray{88.2} & 99.9/\gray{100} & 93.6/\gray{95.1} & 98.7/\gray{100} & 100/\gray{100} & 100/\gray{100} & 100/\gray{100} & 100/\gray{-} & 100±0.00/\gray{100} \\
Tile & 93.7/\gray{99.1} & 93.3/\gray{98.1} & 89.5/\gray{91.6} & 99.8/\gray{99.6} & 97.5/\gray{99.3} & 99.0/\gray{98.9} & 99.3/\gray{99.0} & 99.2/\gray{-} & 100±0.00/\gray{100} \\
Wood & 90.6/\gray{97.7} & 98.4/\gray{99.2} & 93.4/\gray{94.3} & 99.8/\gray{99.1} & 99.2/\gray{99.2} & 99.5/\gray{99.0} & 98.6/\gray{97.9} & 97.2/\gray{-} & 99.6±0.51/\gray{99.2} \\ 
\midrule
Mean & 74.5/\gray{87.7} & 84.2/\gray{95.5} & 81.9/\gray{87.8} & 88.1/\gray{98.0} & 93.4/\gray{98.5} & 97.6/\gray{99.0} & 96.5/\gray{96.6} & 98.0/\gray{-} & \textbf{99.3±0.08}/\textbf{\gray{99.4}} \\
\bottomrule
\end{tabular}
}
\caption{Quantitative results for anomaly detection, evaluated with AUROC metric on MVTec-AD. All methods are evaluated under the unified and \gray{separate} settings.}
\label{tab:detect}
\end{table}

\begin{table}[h!]
\centering
\resizebox{\textwidth}{!}{%
\begin{tabular}{lccccccccc} 
\toprule
Class & US~\cite{bergmann2020uninformed} & PaDiM~\cite{padim} & MKD~\cite{salehi2021multiresolution} & DRAEM~\cite{draem} & RD4AD~\cite{rd4ad} & PatchCore~\cite{patchcore} & UniAD~\cite{uniad} & HVQ-T~\cite{hvq} & CRAD (Ours) \\ 
\midrule
Bottle & 67.9/\gray{97.8} & 96.1/\gray{98.2} & 91.8/\gray{96.3} & 87.6/\gray{99.1} & 97.7/\gray{98.7} & 98.4/\gray{98.6} & 98.1/\gray{98.1} & 98.3/\gray{-} & 98.2±0.10/\gray{98.6} \\
Cable & 78.3/\gray{91.9} & 81.0/\gray{96.7} & 89.3/\gray{82.4} & 71.3/\gray{94.7} & 83.1/\gray{97.4} & 96.7/\gray{98.5} & 97.3/\gray{96.8} & 98.1/\gray{-} & 98.4±0.17/\gray{98.3} \\
Capsule & 85.5/\gray{96.8} & 96.9/\gray{98.6} & 88.3/\gray{95.9} & 50.5/\gray{94.3} & 98.5/\gray{98.7} & 94.8/\gray{98.9} & 98.5/\gray{97.9} & 98.8/\gray{-} & 98.7±0.08/\gray{98.6} \\
Hazelnut & 93.7/\gray{98.2} & 96.3/\gray{98.1} & 91.2/\gray{94.6} & 96.9/\gray{99.7} & 98.7/\gray{98.9} & 98.6/\gray{98.7} & 98.1/\gray{98.8} & 98.8/\gray{-} & 98.5±0.17/\gray{98.9} \\
Metal Nut & 76.6/\gray{97.2} & 84.8/\gray{97.3} & 64.2/\gray{86.4} & 62.2/\gray{99.5} & 94.1/\gray{97.3} & 98.3/\gray{98.4} & 94.8/\gray{95.7} & 96.3/\gray{-} & 97.5±0.36/\gray{97.3} \\
Pill & 80.3/\gray{96.5} & 87.7/\gray{95.7} & 69.7/\gray{89.6} & 94.4/\gray{97.6} & 96.5/\gray{98.2} & 97.3/\gray{97.6} & 95.0/\gray{95.1} & 97.1/\gray{-} & 98.2±0.03/\gray{98.0} \\
Screw & 90.8/\gray{97.4} & 94.1/\gray{98.4} & 92.1/\gray{96.0} & 95.5/\gray{97.6} & 99.4/\gray{99.6} & 98.0/\gray{99.4} & 98.3/\gray{97.4} & 98.9/\gray{-} & 99.3±0.04/\gray{99.2} \\
Toothbrush & 86.9/\gray{97.9} & 95.6/\gray{98.8} & 88.9/\gray{96.1} & 97.7/\gray{98.1} & 99.0/\gray{99.1} & 98.4/\gray{98.7} & 98.4/\gray{97.8} & 98.6/\gray{-} & 98.8±0.04/\gray{98.7} \\
Transistor & 68.3/\gray{73.7} & 92.3/\gray{97.6} & 71.7/\gray{76.5} & 64.5/\gray{90.9} & 86.4/\gray{92.5} & 94.9/\gray{96.4} & 97.9/\gray{98.7} & 97.9/\gray{-} & 98.1±0.14/\gray{98.3} \\
Zipper & 84.2/\gray{95.6} & 94.8/\gray{98.4} & 86.1/\gray{93.9} & 98.3/\gray{98.8} & 98.1/\gray{98.2} & 95.8/\gray{98.9} & 96.8/\gray{96.0} & 97.5/\gray{-} & 97.8±0.06/\gray{97.9} \\
\midrule
Carpet & 88.7/\gray{93.5} & 97.6/\gray{99.0} & 95.5/\gray{95.6} & 98.6/\gray{95.5} & 98.8/\gray{98.9} & 98.9/\gray{99.1} & 98.5/\gray{98.0} & 98.7/\gray{-} & 98.6±0.06/\gray{98.7} \\
Grid & 64.5/\gray{89.9} & 71.0/\gray{97.1} & 82.3/\gray{91.8} & 98.7/\gray{99.7} & 99.2/\gray{99.3} & 96.9/\gray{98.7} & 96.5/\gray{94.6} & 97.0/\gray{-} & 98.0±0.05/\gray{98.0} \\
Leather & 95.4/\gray{97.8} & 84.8/\gray{99.0} & 96.7/\gray{98.1} & 97.3/\gray{98.6} & 99.4/\gray{99.4} & 99.3/\gray{99.3} & 98.8/\gray{98.3} & 98.8/\gray{-} & 98.9±0.07/\gray{99.1} \\
Tile & 82.7/\gray{92.5} & 80.5/\gray{94.1} & 85.3/\gray{82.8} & 98.0/\gray{99.2} & 95.6/\gray{95.6} & 95.9/\gray{95.9} & 91.8/\gray{91.8} & 92.2/\gray{-} & 94.4±0.16/\gray{94.6} \\
Wood & 83.3/\gray{92.1} & 89.1/\gray{94.1} & 80.5/\gray{84.8} & 96.0/\gray{96.4} & 96.0/\gray{95.3} & 94.4/\gray{95.1} & 93.2/\gray{93.4} & 92.4/\gray{-} & 93.8±0.09/\gray{93.8} \\ 
\midrule
Mean & 81.8/\gray{93.9} & 89.5/\gray{97.4} & 84.9/\gray{90.7} & 87.2/\gray{97.3} & 96.0/\gray{97.8} & 97.1/\textbf{\gray{98.1}} & 96.8/\gray{96.6} & 97.3/\gray{-} & \textbf{97.8±0.12}/\gray{97.9} \\
\bottomrule
\end{tabular}
}
\caption{Quantitative results for anomaly localization, evaluated with AUROC metric on MVTec-AD. All methods are evaluated under the unified and \gray{separate} settings.}
\label{tab:local}
\end{table}

\noindent \textbf{Sampled features from the grids (normal vs. abnormal). } 
In addition, we visualize the sampled features from the local and global grids based on the learned coordinates.
\cref{fig:coord} shows that the sampled features (from the local and global grids) with near coordinates share similar characteristics whether the input image is normal or not.
Furthermore, the fused normal representations of both normal and abnormal inputs are reconstructed into normal images.
Specifically, CRAD preserves the fine-grained details of the bottle (normal region), while it reconstructs the corresponding normal state of the anomalous region that has never been encountered during training.
This result demonstrates that the continuous feature space efficiently tackles the two major challenges in discrete feature space: weak generalization and IS.

\noindent\textbf{Decision boundary of multiple classes. }
\cref{fig:global_vis} describes the coordinate distribution using a model trained with global representation.
The images from each class form clusters in the continuous memory space, effectively modeling the decision boundaries between classes.
This implies that the continuous memory can represent well-defined structural features.
Furthermore, the reconstructions of the sampled features at the decision boundary show combined characteristics of near classes, demonstrating the high granularity of continuous features.
Leveraging the advantages of the continuous feature memory, CRAD can model correct decision boundaries in complex multi-class distributions with compact representations, leading to high performance in a unified setting.

\begin{figure}[t]
    \begin{center}
    \includegraphics[width=1.0\linewidth]{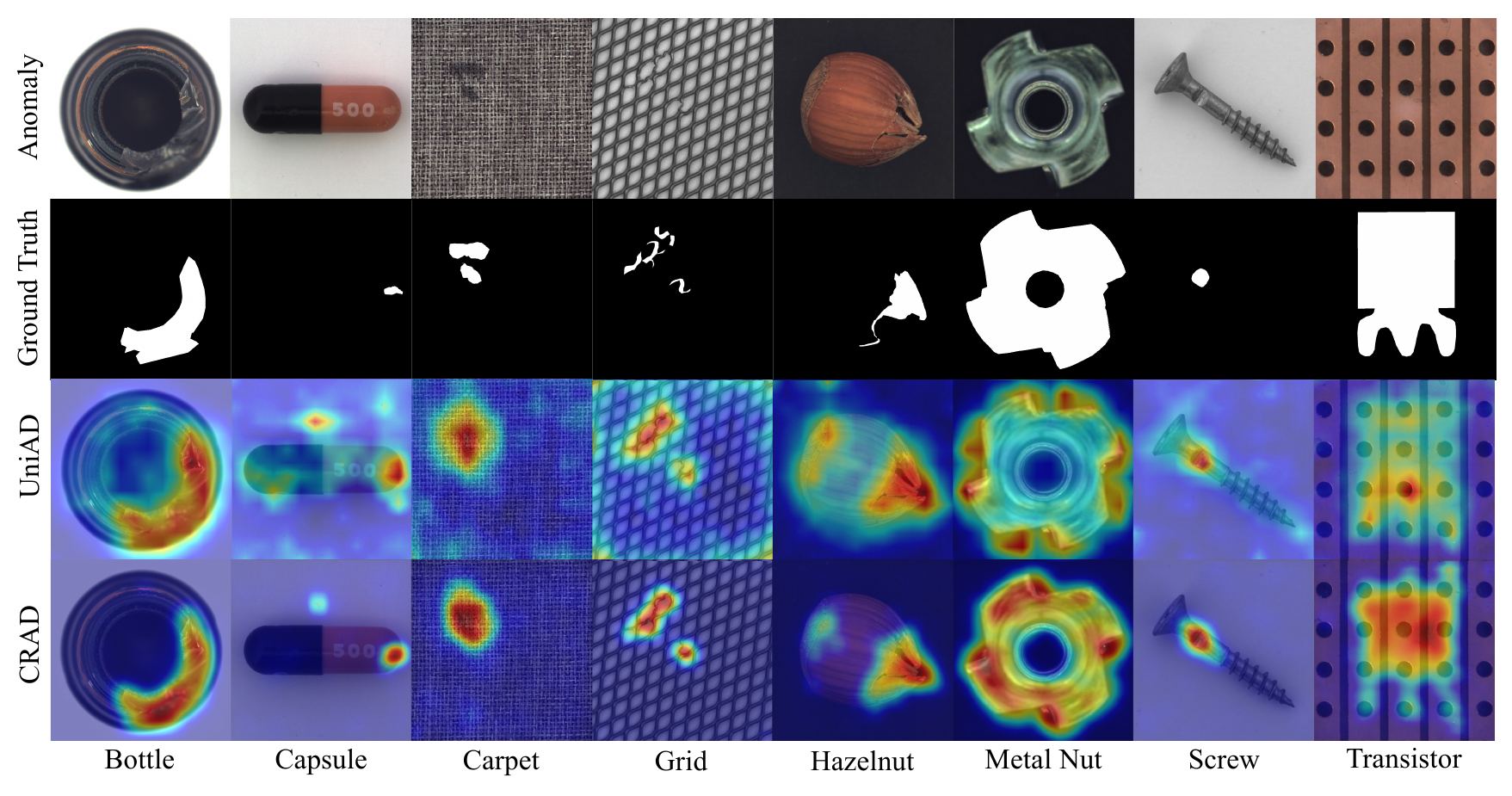}
    \end{center}
    \caption{Qualitative results of CRAD on MVTec AD. Each row of the figure represents anomaly images, corresponding ground truths, results from UniAD, and our results.}
    \label{fig:qual}
\end{figure}
\subsection{Anomaly Detection and Localization}
We evaluate CRAD in comparison with recent state-of-the-art methods in both unified and separate settings, focusing on detection (\cref{tab:detect}) and localization (\cref{tab:local}) performance on MVTec AD.
In the unified setting, the methods not specifically designed for multiple classes exhibit a significant performance drop compared to their performance in the separate setting.
In contrast, UniAD, HVQ-Trans, and CRAD maintain the performances of their separate models in the unified setting.
Among these, CRAD notably outperforms UniAD and HVQ-Trans, achieving state-of-the-art performance in the unified setting.
Specifically, CRAD successfully reduces the error rate of HVQ-Trans from 2.0\% to 0.7\%, bringing a total error reduction of 65.0\%.
For detection, the unified CRAD even outperforms separate models of PatchCore, which is the previous state-of-the-art in single-class AD.
Similarly, for localization, CRAD achieves the best performance in the unified setting and matches PatchCore in a separate setting.
\cref{fig:qual} showcases the qualitative results of UniAD and CRAD in the unified setting, highlighting CRAD's superior prediction quality with fewer noisy areas.
We additionally evaluate CRAD on VisA, which is a more challenging dataset. 
As shown in~\cref{tab:visa}, CRAD outperforms other state-of-the-art methods for detection.

\begin{table*}[t]
\centering
\resizebox{\textwidth}{!}{%
\begin{tabular}{cccccccccc} 
\toprule
\multicolumn{2}{c}{\multirow{2}{*}{Class}} & \multicolumn{4}{c}{Detection} & \multicolumn{4}{c}{Localization} \\ 
\cmidrule(lr){3-6}\cmidrule(lr){7-10}
\multicolumn{2}{c}{} & UniAD~\cite{uniad} & PatchCore~\cite{patchcore} & OmniAL~\cite{omnial} & CRAD (ours) & UniAD~\cite{uniad} & PatchCore~\cite{patchcore} & OmniAL~\cite{omnial} & CRAD (ours) \\ 
\midrule
\multirow{4}{*}{\begin{tabular}[c]{@{}c@{}}Complex \\Structure\end{tabular}} & PCB1 & 94.8/\gray{90.2} & 97.6/\gray{98.5} & 77.7/\gray{96.6} & 96.8/\gray{95.4} & 99.3/\gray{99.2} & 99.7/\gray{99.8} & 97.6/\gray{98.7} & 99.5/\gray{99.5} \\
 & PCB2 & 92.5/\gray{84.2} & 96.7/\gray{97.2} & 81.0/\gray{99.4} & 92.9/\gray{92.7} & 97.6/\gray{96.5} & 98.0/\gray{98.7} & 93.9/\gray{83.2} & 97.6/\gray{97.0} \\
 & PCB3 & 86.6/\gray{90.7} & 97.3/\gray{98.5} & 88.1/\gray{96.9} & 95.2/\gray{96.1} & 98.1/\gray{98.0} & 99.3/\gray{99.4} & 94.7/\gray{98.4} & 98.7/\gray{98.6} \\
 & PCB4 & 99.3/\gray{97.4} & 99.7/\gray{99.7} & 95.3/\gray{97.4} & 99.4/\gray{98.6} & 97.6/\gray{97.2} & 97.7/\gray{98.2} & 97.1/\gray{98.5} & 98.6/\gray{98.4} \\ 
\midrule
\multirow{4}{*}{\begin{tabular}[c]{@{}c@{}}Multiple \\Instances\end{tabular}} 
 & Candle & 97.0/\gray{90.2} & 94.7/\gray{99.4} & 86.8/\gray{85.1} & 96.3/\gray{96.6} & 99.1/\gray{99.0} & 98.3/\gray{99.3} & 95.8/\gray{90.5} & 99.2/\gray{99.2} \\ 
 & Capsules & 70.7/\gray{80.3} & 75.0/\gray{76.3} & 90.6/\gray{87.9} & 90.5/\gray{91.5} & 98.1/\gray{98.5} & 99.1/\gray{99.2} & 99.4/\gray{98.6} & 99.5/\gray{99.5} \\
 & Macaroni1 & 90.4/\gray{90.2} & 94.7/\gray{97.4} & 92.6/\gray{96.9} & 96.6/\gray{96.0} & 99.1/\gray{99.0} & 99.0/\gray{99.7} & 98.6/\gray{98.9} & 99.1/\gray{99.1} \\
 & Macaroni2 & 82.8/\gray{77.4} & 78.6/\gray{76.7} & 75.2/\gray{89.9} & 88.7/\gray{90.4} & 97.7/\gray{97.4} & 96.1/\gray{98.6} & 97.9/\gray{99.1} & 98.8/\gray{99.0} \\
\midrule
\multirow{4}{*}{\begin{tabular}[c]{@{}c@{}}Single\\Instance\end{tabular}} 
 & Cashew & 93.8/\gray{92.9} & 97.3/\gray{97.8} & 88.6/\gray{97.1} & 95.5/\gray{96.4} & 98.9/\gray{99.2} & 98.1/\gray{98.7} & 95.0/\gray{98.9} & 97.4/\gray{98.0} \\
 & Chewinggum & 99.3/\gray{98.3} & 98.5/\gray{98.8} & 96.4/\gray{94.9} & 99.5/\gray{98.9} & 99.1/\gray{98.5} & 98.9/\gray{98.9} & 99.0/\gray{98.7} & 98.3/\gray{98.4} \\
 & Fryum & 88.8/\gray{84.4} & 95.4/\gray{96.0} & 94.6/\gray{97.0} & 94.5/\gray{93.7} & 97.7/\gray{96.7} & 89.8/\gray{92.4} & 92.1/\gray{89.3} & 96.6/\gray{96.3} \\
 & Pipe fryum & 97.0/\gray{91.8} & 99.2/\gray{99.8} & 86.1/\gray{91.4} & 96.6/\gray{98.3} & 99.3/\gray{99.3} & 97.5/\gray{98.9} & 98.2/\gray{99.1} & 99.4/\gray{99.4} \\ 
\midrule
\multicolumn{2}{c}{Mean} & 91.1/\gray{89.0} & 93.7/\gray{94.7} & 87.8/\gray{94.2} & \textbf{95.2/\gray{95.4}} & 98.5/\gray{98.2} & 97.5/\textbf{\gray{98.5}} & 96.6/\gray{96.0} & \textbf{98.6/\gray{98.5}} \\
\bottomrule
\end{tabular}
}
\caption{Quantitative results for anomaly detection and localization, evaluated on VisA. All methods are evaluated under the unified and \gray{separate} settings.}
\label{tab:visa}
\end{table*}

\subsection{Ablation Study}\label{sec:ablation}
\begin{minipage}[b]{0.6\textwidth}
We conducted an ablation study on CRAD to assess the impact of individual proposals. 
\cref{tab:abl} shows that our key contribution is the normal representation from both local and global contexts, which independently yields comparable performance.
Notably, a model with only local representation outperforms UniAD and HVQ-
\end{minipage}\hfill
\begin{minipage}[b]{0.38\textwidth}
\captionof{table}{Ablation studies in the unified setting using MVTec AD.}
\centering
\resizebox{1.0\textwidth}{!}{
\begin{tabular}{cccccc} 
        \toprule
        Local & Global & Refine & Jitter & Detect & Localize \\ 
        \cmidrule(lr){1-4}\cmidrule(lr){5-6}
        & \checkmark &  &  & 92.3 & 95.7 \\
        \checkmark &  &  &  & 98.6 & 97.5 \\
        \checkmark & \checkmark &  &  & 98.8 & 97.7 \\ 
        \cmidrule(lr){1-4}\cmidrule(lr){5-6}
        \checkmark & \checkmark & \checkmark &  & 99.1 & 97.8 \\
        \checkmark & \checkmark & \checkmark & \checkmark & 99.3 & 97.8 \\
        \bottomrule
    \end{tabular}}
\label{tab:abl}
\end{minipage}
Trans. The integration of both representations achieves improved performance, with additional gains from feature refinement and coordinate jittering.

\section{Conclusion}
In this work, we have proposed a novel anomaly detection architecture, CRAD, which represents normal features in the continuous memory, unlike prior approaches limited to discrete feature space.
CRAD successfully represents local as well as global features in the continuous space while overcoming the limitations of existing methods, such as weak generalization, identity shortcut, and high computational/parameter complexity.
Through extensive experiments, we have demonstrated the effectiveness of CRAD qualitatively and quantitatively.
Although CRAD demonstrates its superior generalization capability compared to existing methods, we found a limitation that this cannot be the case with extremely limited data (i.e., 1- or zero-shot), more discussed in the supplementary materials.
We believe that it can be further addressed and our work paves the way for future advancements in anomaly detection.

\section*{Acknowledgements}
This work was supported in part by the Institute of Information and Communications Technology Planning and Evaluation (IITP) funded by the Korea Government (MSIT) under Grant RS-2021-II212068 (Artificial Intelligence Innovation Hub) and Grant RS-2022-II220688 (AI Platform to Fully Adapt and Reflect Privacy-Policy Changes); in part by the Culture, Sports, and Tourism R\&D Program through the Korea Creative Content Agency funded by the Ministry of Culture, Sports and Tourism in 2024 under Grant RS-2024-00348469 (Research on neural watermark technology for copyright protection of generative AI 3D content); and in part by the SEMES-Sungkyunkwan University collaboration funded by SEMES.
%
%
\bibliographystyle{splncs04}
\bibliography{main}

\appendix
\section{Implementation Details}
\label{app_imp}
We implemented CRAD in the PyTorch framework, and we used the NVIDIA A5000 GPU for all evaluations.

\noindent\textbf{Training. }
We trained our models for 50 epochs, thrice with different seeds (0,1,2), with a batch size of 64.
The learning rate is initially set to $1\times10^{-3}$ and $1\times10^{-1}$ for neural networks and grids, respectively, which are reduced by a factor of 0.1 once at the 40 epoch. We used AdamW optimizer~\cite{adamw} with weight decay $1\times10^{-2}$.

\noindent\textbf{Model configurations.}
To create the input feature $x$, we utilized the third and fourth stage feature maps of EfficientNet-b4~\cite{effinet}, which are resized to $28 \times 28$ and then concatenated (channel $C=216$).
We configured both the dimensions of the local ($C_l$) and global coordinates ($C_g$) to 2. 
Additionally, we set the grid resolution of the local ($R_l$) and global ($R_g$) representations to 8 and 4, respectively. 
In terms of feature refinement, we set the parameters $\lambda_1$, $\lambda_2$, and $k$ to be 0.3, 0.7, and 10.
For coordinate jittering, we applied Gaussian noise $N(0,1)$ scaled by a factor of 0.05 to 50\% of pixel-wise features.
We designed the convolutional network $\psi$ with 11 MobileNetV3~\cite{resnet} blocks that maintain the number of channels and the spatial resolution.
Unless otherwise noted, the model performances were measured in a unified setting.

\section{Additional Experiments}
We conducted more extensive experiments to validate the effectiveness in terms of generalization ability.

\begin{table}
\centering
\begin{tabular}{cccccc} 
\toprule
Method & FLOPs & \#Params & Epoch & Detect & Localize \\ 
\cmidrule(lr){1-1}\cmidrule(lr){2-4}\cmidrule(lr){5-6}
UniAD & 6.5G & 7.5M & 1000 & 96.5 & 96.8 \\
CRAD-S & 3.7G & 4.5M & 50 & 99.0 & 97.8 \\
CRAD & 13.5G & 14.6M & 50 & 99.3 & 97.8 \\
\bottomrule
\end{tabular}
\caption{The evaluation of complexity, model size, and training duration with the AUROC performance on MVTec AD, where \#Params denotes the number of parameters.}
\label{tab:eff}
\end{table}

\subsection{Computational and Parameter Efficiency}
We evaluated the model size and computational complexity of CRAD.
As shown in~\cref{tab:eff}, CRAD requires reasonable storage and computation resources.
However, for a fair comparison with UniAD, we downsized the convolutional networks while maintaining the grid configurations, coined as CRAD-S.
CRAD-S exhibits superior performance with reduced memory and computation requirements, demonstrating both effectiveness and efficiency.
Furthermore, CRAD has another strong point: fast training time enabled by grids.
The outstanding performance is achieved only with $1/20$ training duration compared to UniAD.

\subsection{Generalization Ability}
\label{sec:gen}
We have shown the effectiveness of the continuous feature space when sufficient training data is available.
However, in this subsection, we suppose a more challenging circumstance where the quantity of normal training data is significantly limited.
When the normal images are not sufficient, models often suffer from learning the representative characteristics of the class.
For instance, in discrete feature space, if an input feature significantly deviates from the normal features in the space, the model struggles to distinguish whether the distance stems from unseen normal or abnormal features since it has few normal features to be referenced.
We compared CRAD with other methods by training each model with few-shot images during 100 epochs.
As shown in~\cref{tab:generalization}, the methods with discrete feature space show limited performance under these constraints.
On the other hand, CRAD allows continuous normal representation by interpolating the learned normal features in spite of the scarcity, resulting in superior performance under few-shot settings.
This demonstrates the generalization ability of the continuous space even in a more challenging scenario.
Despite outperforming the other baselines, CRAD's performance under the extremely scarce dataset (i.e., 1-shot scenario) leaves room for improvement and we believe is not entirely satisfactory for practical uses.
We recognize this as a limitation of CRAD, which will be discussed in~\cref{sec:lim}.

\begin{table*}[t]
\centering
\begin{tabular}{ccccccc} 
\toprule
\multirow{2}{*}[-0.3em]{Setting} & \multicolumn{3}{c}{Detection} & \multicolumn{3}{c}{Localization} \\ \cmidrule(lr){2-4}\cmidrule(lr){5-7}
 & PatchCore & UniAD & CRAD & PatchCore & UniAD & CRAD \\ 
\midrule
1-shot & 82.7 & 80.3 & 83.3 & 91.8 & 91.2 & 92.2 \\
2-shot & 87.5 & 82.2 & 88.2 & 93.8 & 92.4 & 95.1 \\
4-shot & 89.3 & 83.6 & 92.4 & 94.6 & 93.1 & 96.2 \\
\bottomrule
\end{tabular}
\caption{The performance evaluation with few-shot images of each class in MVTec AD. The performances are measured in AUROC.}
\label{tab:generalization}
\end{table*}

\begin{table}[t]
\centering
\begin{tabular}{cccc} 
\toprule
Local & Global & Detect & Localize \\ 
\cmidrule(lr){1-2}\cmidrule(lr){3-4}
1D (64) & 2D (4) & 99.2 & 97.7 \\
3D (8) & 2D (4) & 99.3 & 97.7 \\ 
\cmidrule(lr){1-2}\cmidrule(lr){3-4}
2D (8) & 1D (16) & 99.2 & 97.7 \\
2D (8) & 3D (4) & 99.1 & 97.8 \\ 
\cmidrule(lr){1-2}\cmidrule(lr){3-4}
\textbf{2D (8)} & \textbf{2D (4)} & 99.3 & 97.8 \\
\bottomrule
\end{tabular}
\caption{Ablation study on the grid dimension ($C_l$ and $C_g$) of local and global representations. The values in the parenthesis denote the grid resolution ($R_l$ and $R_g$) of each dimension. For instance, \textbf{2D (4)} in the first row denotes that we set the grid dimension to 2 and each dimension's resolution to 4. The performances are measured in AUROC on MVTec AD. CRAD's setting is highlighted in \textbf{bold}.}  
\label{tab:res}
\end{table}

\subsection{More Ablation Studies}
\noindent\textbf{Grid dimension.}
We implemented CRAD with different grid dimensions (equivalent to the dimension of coordinates).
As shown in~\cref{tab:res}, although the higher dimension of local and global representation slightly increases the detection and localization performance respectively, two representations both with 2-dimension are well balanced, resulting in the optimal performance.
However, the results indicate CRAD is not sensitive to grid dimensions, which can be beneficial in a practical scenario.

\noindent\textbf{Feature refinement.}
We conducted an ablation study on the feature refinement of CRAD, which employs both Mean Squared Error (MSE) and cosine similarity. 
As shown in~\cref{tab:refi}, MSE and cosine similarity show comparable performance when they are applied individually.
By combining the two metrics, CRAD achieves the best performance.

\begin{table}[t]
\centering
\begin{tabular}{@{}ccc@{}} 
\toprule
Refinement & \begin{tabular}[c]{@{}c@{}}Detect\end{tabular} & \begin{tabular}[c]{@{}c@{}}Localize\end{tabular} \\ 
\cmidrule(lr){1-1}\cmidrule(lr){2-3}
MSE & 99.0 & 97.6 \\
Cosine similarity & 98.7 & 97.8 \\
Combined (CRAD) & 99.3 & 97.8 \\
\bottomrule
\end{tabular}
\caption{Ablation study on the feature refinement using MVTec AD. The performances are measured in AUROC.}
\label{tab:refi}
\end{table}

\subsection{Additional Dataset}
We conducted an additional experiment using Real-IAD~\cite{realiad} (single view), a recently introduced and more challenging 30-class AD dataset.
As shown in \cref{tab:iad}, CRAD outperforms UniAD by a large margin. 

\begin{table}[]
\centering
\begin{tabular}{lcc}
\toprule
 Method & Det. (AUROC) & Loc. (AUPRO) \\\midrule
 UniAD & 82.9 & 86.1 \\
 CRAD & \textbf{88.6} & \textbf{87.5} \\\bottomrule
\end{tabular}
\caption{Performance evaluation using Real-IAD dataset in the unified setting.}
\label{tab:iad}
\end{table}

\subsection{More Qualitative Results}
\begin{figure*}[t]
    \begin{center}
    \includegraphics[width=1.0\linewidth]{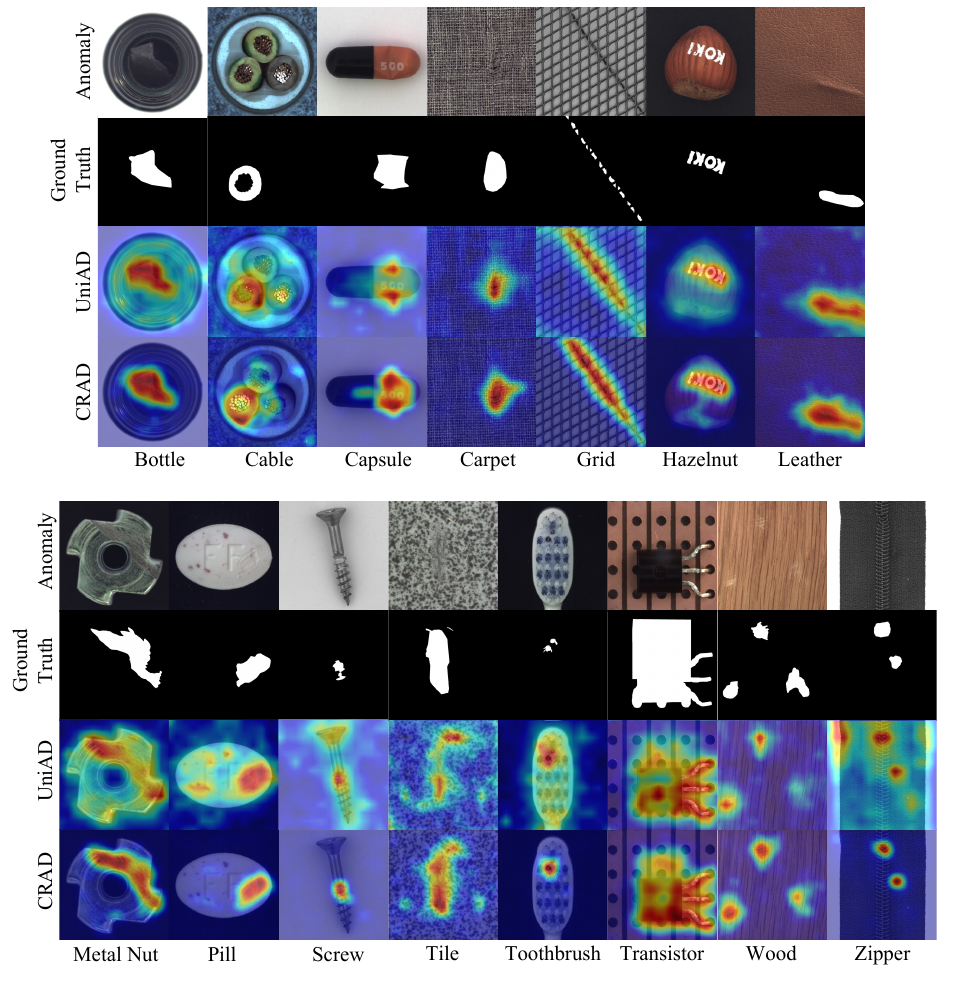}
    \end{center}
    \caption{Qualitative results of CRAD for each class of MVTec AD. Each row represents anomaly images, ground truths, prediction of UniAD, and prediction of CRAD.}
\label{fig:supp_qual}
\end{figure*}

We provide more qualitative results of CRAD in \cref{fig:supp_qual}.

\section{Limitation}\label{sec:lim}
Although CRAD demonstrates its superior generalization capabilities compared to existing methods, as shown in~\cref{sec:gen}, it struggles when dealing with novel features. 
For instance, CRAD can generalize a sparse set of features in a class (i.e., few-shot) by interpolating within their feature space. 
However, this cannot be the case with extremely limited or no data (i.e., 1-shot or zero-shot). 
As this problem has not been fully explored in this field, we believe it can be further addressed in the near future.

\end{document}